\documentclass[letterpaper]{article}

\usepackage[draft]{aaai2026}
\usepackage{times}
\usepackage{helvet}
\usepackage{courier}
\usepackage[hyphens]{url}
\usepackage{graphicx}
\usepackage{natbib}
\usepackage{caption}
\usepackage{algorithm}
\usepackage{algorithmic}
\usepackage{enumitem}
\usepackage{amsmath}
\usepackage{amssymb}
\usepackage{booktabs}
\usepackage{multirow}

\title{ChainPrune: Evaluating and Reducing Redundancy in\\Long Chain-of-Thought Reasoning}

\author{
Weihang Pan\textsuperscript{\rm 1,\textdagger},
Zhengxu Yu\textsuperscript{\rm 2},
Yuxiang Zhang\textsuperscript{\rm 1},
Wenzhi Li\textsuperscript{\rm 1},\\
Zhongming Jin\textsuperscript{\rm 2},
Binbin Lin\textsuperscript{\rm 1,*},
Xiaofei He\textsuperscript{\rm 1},
Jieping Ye\textsuperscript{\rm 2}
}

\affiliations{
\textsuperscript{\rm 1}Zhejiang University\\
\textsuperscript{\rm 2}Alibaba Group\\
panweihang@zju.edu.cn, binbinlin@zju.edu.cn
}

\begin{document}

\maketitle
\begingroup
\renewcommand{\thefootnote}{}
\footnotetext{%
\setlength{\parindent}{0pt}%
\hspace*{-1.8em}\makebox[1.2em][l]{\textsuperscript{*}}Corresponding author.\par
\hangindent=1.2em
\hangafter=1
\noindent\makebox[1.2em][l]{\textsuperscript{\textdagger}}This work was completed during an internship at Alibaba Group in May 2025.}
\endgroup

\begin{abstract}
Chain-of-Thought (CoT) reasoning has significantly enhanced the multi-step problem-solving capabilities of large language models (LLMs) by introducing explicit intermediate reasoning. However, advanced Large Reasoning Models (LRMs) often exhibit overthinking behaviors, including excessively long reasoning steps, redundant steps, and high computational overhead. Existing token-length reward strategies aim to promote concise outputs, but often result in pseudo-conciseness, where token count is reduced, yet redundant reasoning persists, leading to longer and less structurally efficient chains. To address these limitations, we propose \textbf{ChainPrune}, a novel reasoning path semantic structural optimization method to efficiently and controllably synthesize self-generated high-quality training data. We initially consolidate self-generated reasoning paths into a tree-based structure, followed by a multi-criteria dominant path selection process for preference data construction that formulates shallow reasoning trajectories while preserving essential reasoning steps. To further enhance the quality of reasoning, we incorporate a DPO-based preference learning method combined with supervised loss, effectively mitigating false reward suppression. This innovative integration significantly enhances both the efficiency and effectiveness of our reasoning framework.
Comprehensive experimental results demonstrate significant reductions in step length and computational overhead, while maintaining or even enhancing accuracy.
\end{abstract}

\section{Introduction}
 The emergence of Chain-of-Thought (CoT) \cite{wei2022chain} reasoning has signaled a paradigm shift in large language models (LLMs) by facilitating multi-step problem-solving through the articulation of explicit intermediate reasoning steps. Initial CoT methods revealed that breaking down complex tasks into structured steps---mirroring human deliberation---can significantly improve performance across various domains, including mathematics and symbolic reasoning. 

 Recent advances in Large Reasoning Models (LRMs) \cite{xu2025towards}, such as OpenAI's O1 \cite{jaech2024openai} and DeepSeek-R1 \cite{guo2025deepseek}, have systematized the Chain-of-Thought (CoT) process through differentiated reinforcement learning paradigms, incorporating innovative architectures like Monte Carlo Tree Search (MCTS) \cite{browne2012survey, coulom2006efficient} and pure reinforcement learning strategies. These o1-like models emulate human-like "slow thinking" \cite{li2025system} by exploring multiple solution strategies, engaging in self-reflection, and iteratively correcting errors \cite{liang2024internal}. However, this deliberative approach gives rise to the \textbf{overthinking phenomenon} \cite{chen2024not, team2025kimi}, characterized by exponential proliferation of reasoning steps, excessive verbosity, and redundancy. Such inefficiencies result in significant computational overhead and error accumulation, ultimately impeding the practical deployment of these models. Addressing these challenges is crucial for enhancing the utility and applicability of LRMs in real-world scenarios.

\begin{figure}
    \centering
    \includegraphics[width=\linewidth]{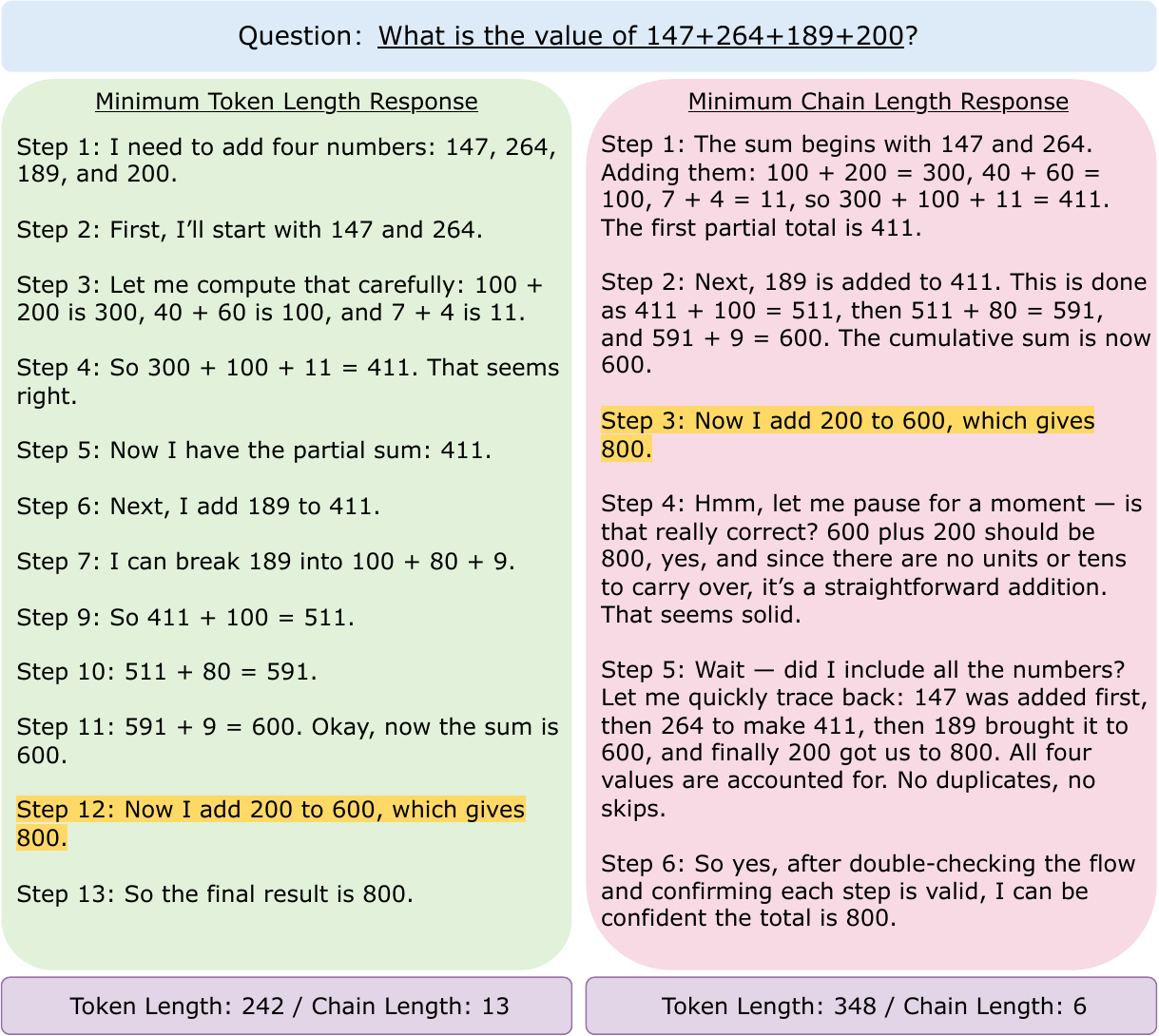}
    \caption{Comparison between a Minimum Token Length Response and a Minimum Chain Length Response for the same addition problem.}
    \label{fig:1}
\end{figure}

To mitigate this, researchers have pursued efficient reasoning techniques to shorten reasoning sequences while preserving accuracy \cite{han2024token, hao2024training, luo2025o1, ma2025cot, yeo2025demystifying}. Recent guide-based methods \cite{liao2025reward} attempt to address this by framing the reasoning process as a tree search problem, then steering the search toward concise outputs through manual rules or step-by-step supervision. However, these approaches introduce reward bias that propagates reasoning errors. Furthermore, several methods \cite{luo2025o1, team2025kimi, shen2025dast, arora2025training} introduce token-length reward mechanisms during RL training to guide concise outputs.
Yet, such token--length--only optimization often leads to what we call pseudo-conciseness: while the token count is reduced, the reasoning path is fragmented into many micro-steps, inflating chain length and reducing structural efficiency. Figure~\ref{fig:1} illustrates this phenomenon with a simple arithmetic example: the Minimum Token Length Response minimizes tokens by using terse phrases per step, but its chain length becomes unnecessarily long. In contrast, the Minimum Chain Length Response consolidates the reasoning into fewer, semantically richer steps, achieving shorter chain length but not necessarily minimizing tokens. The ideal scenario---and the target of our work---is to reduce both token usage and chain length simultaneously.

To this end, we propose ChainPrune, a novel low-cost reasoning path semantic structural optimization method that synthesizes high-quality training data by merging semantically equivalent reasoning steps across multiple sampled paths. By consolidating redundant reasoning into a compact, logically complete chain, ChainPrune produces reasoning paths that are shorter in both tokens and steps, without sacrificing correctness or completeness. This joint optimization ensures that our models avoid pseudo-conciseness and instead generate genuinely efficient reasoning sequences.
We initially consolidate self-generated reasoning paths into a tree-based structure, with dynamic semantic node merging to prune redundancy while preserving critical reasoning points. Subsequently, we introduce a multi-criteria dominant path selection mechanism to construct the preference dataset, which jointly optimizes chain and token efficiency. For chosen response selection, we prioritize Pareto-dominant reasoning paths when available; otherwise, we favor shorter responses among non-dominated candidates. Rejected responses are required to exhibit substantially longer reasoning steps or incorrect logic to ensure meaningful comparison.
We then train a variant of DPO that includes a negative log-likelihood (NLL) loss term for the winning pairs, which proves crucial for enforcing both correctness and efficiency in reasoning quality.

Our main contributions are as follows:
\begin{itemize}[leftmargin=10pt]
    \item We reveal that current token-length optimization methods suffer from pseudo-conciseness, where token count is reduced but reasoning paths remain redundant. We propose ChainPrune, a low-cost compression method based on semantic structural similarity, which merges semantically equivalent nodes in sampled reasoning paths to reduce redundant steps while preserving core logic.
    \item We design a novel preference data construction pipeline that dynamically searches for Pareto-optimal paths from merged reasoning trees, automatically generating high-quality training data without manual annotation. Additionally, our experiments compare three preference learning strategies (DPO \cite{rafailov2023direct}, SimPO \cite{meng2024simpo}, and DPO + NLL Loss \cite{pang2024iterative}), showing that the DPO + NLL Loss objective significantly shortens reasoning steps while maintaining reasoning quality.
    \item We conduct extensive experiments across 5 reasoning benchmarks, showing \textbf{ChainPrune} improves reasoning efficiency by 26.8\% chain lengths and 28.1\% token lengths reduction without accuracy drop. To holistically evaluate reasoning quality, we design an \textbf{LLM-as-a-Judge} framework with human verification, analyzing faulty reasoning, reflection mechanisms, and step efficiency. The results demonstrate significant reductions of 57.8\% in faulty steps, 62.1\% in invalid reflections, and 65.9\% in redundant steps compared to the base model. This reveals our method's superiority in maintaining coherent logic while minimizing redundant deliberation. 
\end{itemize}

\begin{figure*}
    \centering
    \includegraphics[width=0.80\linewidth]{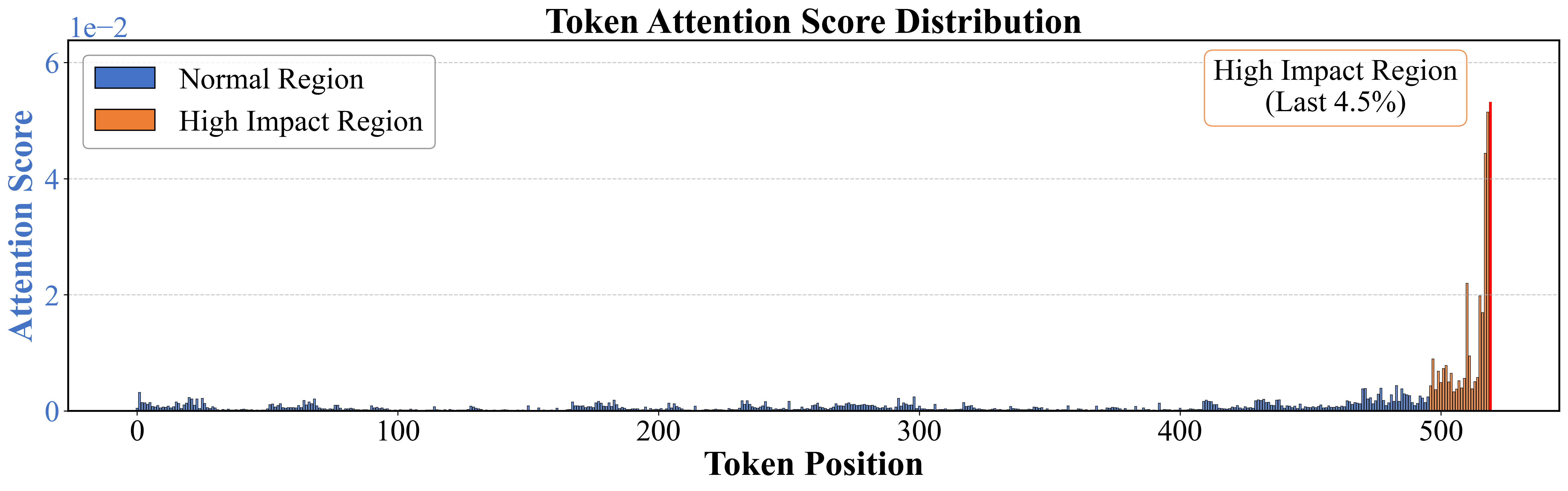}
    \caption{Attention weight distribution following semantically equivalent steps, showing localized focus on the most recent tokens (x-axis: Token Position, y-axis: Attention Score). This supports our tree-merging approach: merging equivalent nodes preserves the conditional distribution of subsequent tokens, enabling both token and step reduction without loss of correctness.}
    \label{fig:2}
\end{figure*}

\section{Formal Analysis of Reasoning}
\subsection{Formal Modeling of Autoregressive Inference}
The inference process of large language models (LLMs) can be formalized as an \emph{autoregressive generation paradigm}:  
Given an initial context $\boldsymbol{x}_0 = (x_{01}, \dots, x_{0k}) \in \mathcal{A}^k$, the model generates tokens sequentially as:
\begin{equation}
    x_{i+1} \sim \mathcal{C}_{i+1} = f(\boldsymbol{x}_0 \oplus x_1 \oplus \cdots \oplus x_i),
\end{equation}
where $\oplus$ denotes sequence concatenation, and $\mathcal{C}_{i+1}$ is a probability distribution over vocabulary $\mathcal{A}$. This process can be naturally viewed as a \textbf{tree search problem}: The root node is the initial input $\boldsymbol{x}_0$, each branch represents a candidate token expansion $a \in \mathcal{A}$, and the resulting inference tree $\mathcal{T}$ grows exponentially with depth ($\mathcal{O}(b^d)$, $b=|\mathcal{A}|$).

Recent reasoning-focused LLMs, such as OpenAI-O1 \cite{jaech2024openai} and DeepSeek-R1 \cite{guo2025deepseek}, extend this paradigm from single-step prediction to structured chain-of-thought generation.  
Implicit CoT methods construct reasoning chains without explicit intermediate supervision, whereas explicit-format approaches separate reasoning from final answers via semantic tags (e.g., $<$think$>$, $<$answer$>$) and apply format-constrained rewards.  
State-of-the-art methods further integrate self-verification (e.g., code execution, symbolic checks) to detect and correct reasoning errors within differentiable feedback loops.
While these advances improve reasoning quality, they also reveal inefficiencies in how reasoning paths are generated and selected---particularly in reinforcement learning--optimized models---leading to the challenges described next.

\subsection{Challenges in RL-Optimized Reasoning Paths}
\label{challenges in reasoning paths}
Although reinforcement learning (RL) fine-tuning enhances reasoning ability, our analysis shows that reward designs often prioritize \emph{token-level conciseness} while neglecting the \emph{structural length} of reasoning chains. This bias produces what we call \textbf{pseudo-conciseness}: responses that minimize tokens by using short phrases per step, yet fragment the logic into many steps, resulting in long and inefficient chains.

This issue directly connects to the phenomenon illustrated in Figure~\ref{fig:1}: for the same arithmetic question, a \emph{minimum-token-length} path can have far more reasoning steps than a more compact chain. However, most existing preference-learning pipelines (e.g., DPO) still select ``chosen'' responses solely by minimal token length among correct outputs. Our statistical study confirms the misalignment---only 33.95\% of the shortest-token correct paths are also the shortest in steps (see Appendix for detailed analysis).

To address this gap, we define a \textbf{path domination} criterion:  
Given two correct paths $\tau_1$ and $\tau_2$, $\tau_1$ \emph{dominates} $\tau_2$ ($\tau_1 < \tau_2$) if $\tau_1$ has \emph{both} fewer reasoning steps and shorter token length.  
A \textbf{Pareto-dominant} path dominates all others; when no such path exists, the set of \emph{non-dominated} candidates forms the Pareto frontier.  
This dual-objective selection principle directly aligns with our goal: identify reasoning chains that are concise in both \emph{expression} and \emph{structure}.

\subsection{Cross-Path Semantic Reusability in Tree Search}
Beyond selection criteria, our investigation of sampled reasoning paths reveals another key observation: \textbf{semantically equivalent intermediate steps occur frequently across different paths}. These equivalent steps appear in both ``chosen'' and ``rejected'' candidates and often emerge at similar structural positions.

This recurrence implies an opportunity for \emph{semantic composability} in the tree search space.  
If two reasoning paths share a semantically equivalent node ($\text{node}_i$), their subsequent continuations ($\text{node}_{i+1}$, $\text{node}_{i+2}$, ...) are largely determined by the \emph{local} context of that node.  
Figure~\ref{fig:2} shows that, following semantically equivalent nodes, the model's attention focuses on the most recent tokens, indicating that merging at this point preserves the conditional distribution.

We exploit this property via a \textbf{tree-merging} strategy:  
Merge semantically equivalent nodes across sampled paths and reuse the best-performing continuations from these nodes.  
This process eliminates duplicated segments, producing composite paths that are \emph{shorter in steps} and \emph{shorter in tokens}, without compromising correctness.

This cross-path semantic reusability is a central mechanism in our proposed \textbf{ChainPrune} framework.  
By integrating it with the Pareto-dominance criterion, ChainPrune transforms diverse, verbose sampled outputs into compact, high-quality reasoning chains---achieving simultaneous token and step efficiency while preserving logical integrity.

\begin{figure*}
    \centering
    \includegraphics[width=0.90\linewidth]{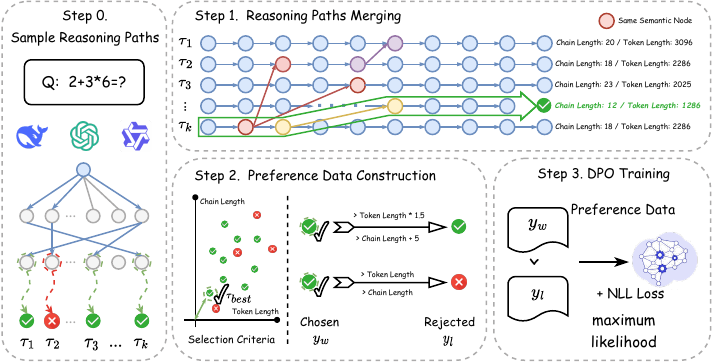}
    \caption{The pipeline of ChainPrune, featuring three synergistic stages: (1) Reasoning Paths Merging with dynamic semantic node pruning, (2) Multi-criteria dominant path selection and Preference Data Construction, (3) Direct preference learning with NLL Loss.}
    \label{chainprune_pipeline}
\end{figure*}

\section{Methodology}
\label{sec:method}
In this section, we present \textbf{ChainPrune}, a framework for generating concise and reliable reasoning chains by pruning redundancy in sampled paths and learning from structured preference signals. The core novelty lies in integrating \emph{semantic path merging} with \emph{dual-objective path selection} (optimizing both token length and reasoning step length) and \emph{enhanced preference optimization}. As illustrated in Figure~\ref{chainprune_pipeline}, ChainPrune operates in three synergistic stages:  
(1) \textbf{Reasoning Paths Merging}, which consolidates multiple sampled reasoning paths into a unified tree structure via semantic node merging, while pruning redundant steps and preserving key reasoning points;  
(2) \textbf{Multi-Criteria Dominant Path Selection and Preference Dataset Construction}, which applies Pareto-dominance criteria to jointly optimize chain and token efficiency;  
(3) \textbf{Preference Optimization}, which fine-tunes the model with an enhanced DPO objective incorporating negative log-likelihood (NLL) regularization.

\subsection{Reasoning Paths Merging}
Given $K$ candidate reasoning paths $\{\tau_1,\dots,\tau_K\}$ generated by the base LLM, this stage constructs a compact reasoning tree that preserves all essential trajectories while removing redundant steps. Each reasoning step is represented as an embedding vector, and merging proceeds as follows.

\textbf{(1) Semantic Node Matching.}
We match a new step $u$ to an existing node $v$ in the tree based on \emph{cosine similarity}:
\begin{equation}
    \text{Sim}(u,v) = \frac{\langle \mathbf{e}_u, \mathbf{e}_v\rangle}{\|\mathbf{e}_u\|\,\|\mathbf{e}_v\|}
\end{equation}
where $\mathbf{e}_u$ and $\mathbf{e}_v$ are the embeddings of steps $u$ and $v$. If $\text{Sim}(u,v) \geq \theta_{\text{sim}}$, the step is considered semantically equivalent to $v$ and is merged; otherwise, a new branch is created. Here $\theta_{\text{sim}}$ is the similarity threshold controlling merge strictness, distinct from any path-related thresholds used elsewhere in our framework.

\textbf{(2) Avoiding Incorrect Merges.}
Semantic similarity alone may mistakenly identify nodes that are lexically similar but logically different as equivalent. 
We therefore add a \emph{token-level entropy criterion} to verify merge candidates. 
Given two candidate nodes $u$ and $v$, we simulate replacing $u$ with $v$ in its original context and compute the Shannon entropy of the token probability distribution over the entire reasoning chain before and after the replacement:
\begin{equation}
    H = -\sum_{t \in \mathcal{A}} P(t) \log P(t),
\end{equation}
where $P(t)$ is the model's predicted token probability and $\mathcal{A}$ is the vocabulary.
Let $H_{\text{before}}$ and $H_{\text{after}}$ be the entropy before and after replacement.
The merge is accepted only if $\Delta H = H_{\text{after}} - H_{\text{before}} \le \epsilon$, meaning predictive uncertainty does not increase. 
This two-stage filtering---combining cosine similarity with entropy change---ensures merges remain semantically and logically consistent.


\subsection{Multi-criteria Dominant Path Selection and Preference Data Construction}
\label{data_construction}
From the merged reasoning tree, we construct preference datasets via a two-phase selection process.

\textbf{(1) Pareto-Dominant Selection phase.}
For each candidate responses $\{\tau_1,\dots,\tau_K\}$, we define the efficiency score:
\begin{equation}
    \mathcal{E}(\tau_j) = \sqrt{\ell_{\text{token}}^2(\tau_j) + \ell_{\text{step}}^2(\tau_j)}
\end{equation}
where $\ell_{\text{token}}(\tau_j)$ and $\ell_{\text{step}}(\tau_j)$ denote token length and chain length, respectively. The preferred answer $\tau^+$ is:
\begin{equation}
    \tau^+ = \underset{\tau_j \in \mathcal{\tau}_{\text{correct}}}{\arg\min}\ \mathcal{E}(\tau_j)
\end{equation}
with $\mathcal{\tau}_{\text{correct}}$ being the set of responses matching the ground truth. If a unique Pareto-dominant path exists (shortest in both token and step length), it is chosen; otherwise, we select the shortest-token path among non-dominated Pareto-front candidates.

\textbf{(2) Threshold-based Rejection phase.}
We filter out low-quality responses $\tau^-$ using two rules:
\begin{itemize}[leftmargin=10pt]
    \item \textbf{Overlong but Correct}: Correct responses that are significantly longer than the chosen $\tau^+$ in both dimensions:  
    $\ell_{\text{step}}(\tau_j) \geq \ell_{\text{step}}(\tau^+) + 5$ \ \textit{and} \ $\ell_{\text{token}}(\tau_j) \geq 1.5\,\ell_{\text{token}}(\tau^+)$.
    \item \textbf{Longer and Incorrect}: Incorrect responses that exceed $\tau^+$ in both step and token length:  
    $\ell_{\text{step}}(\tau_j) > \ell_{\text{step}}(\tau^+)$ \ \textit{and} \ $\ell_{\text{token}}(\tau_j) > \ell_{\text{token}}(\tau^+)$.
\end{itemize}
This ensures that rejected samples are unambiguously inferior in both efficiency and correctness.

\begin{table*}[t]
    \centering
    \small
    \setlength{\tabcolsep}{1mm}
    \begin{tabular}{lcccccccccccc}
    \toprule
    \multirow{2}{*}{\textbf{Methods}} & \multicolumn{3}{c}{\textbf{AIME24}} & \multicolumn{3}{c}{\textbf{AIME25}} & \multicolumn{3}{c}{\textbf{AMC23}} & \multicolumn{3}{c}{\textbf{LiveCodeBench}} \\
    \cmidrule(lr){2-4} \cmidrule(lr){5-7} \cmidrule(lr){8-10} \cmidrule(lr){11-13}
    & ACC & Tokens & Chains & ACC & Tokens & Chains & ACC & Tokens & Chains & ACC & Tokens & Chains \\
    \midrule
    \multicolumn{13}{c}{\textit{DeepSeek-R1-Distilled-Qwen-7B}} \\
    Base Model & \underline{0.5479} & 7346 & 253 & \underline{0.4229} & 6892 & 240 & 0.9047 & 5407 & 189 & 0.3127 & 3961 & 166 \\
    SFT & 0.5437 & 6296 & \textbf{199} & 0.3979 & 5570 & \underline{192} & \textbf{0.9219} & \underline{3824} & \textbf{120} & \underline{0.3344} & 3494 & 143 \\
    Kimi-1.5 & 0.5125 & \underline{6144} & 306 & 0.3833 & \textbf{5292} & 254 & 0.8688 & 4265 & 237 & 0.1703 & \textbf{1953} & \textbf{57} \\
    DAST & 0.5330 & 6337 & - & - & - & - & - & - & - & - & - & - \\
    DAST (reproduce) & 0.5375 & 6909 & 244 & 0.4062 & 6595 & 242 & 0.8891 & 4774 & 160 & 0.3220 & 3370 & 136 \\ 
    \textbf{ChainPrune} & \textbf{0.5833} & \textbf{5688} & \underline{225} & \textbf{0.4250} & \textbf{5351} & \textbf{189} & \underline{0.9125} & \textbf{3771} & \underline{122} & \textbf{0.3498} & \underline{3048} & \underline{116} \\
    \midrule
    \multicolumn{13}{c}{\textit{DeepSeek-R1-Distilled-Qwen-1.5B}} \\
    Base Model & \underline{0.3104} & 7015 & 225 & 0.2167 & 5248 & 158 & 0.7016 & 5316 & 167 & \underline{0.1455} & 2661 & 90 \\
    SFT & 0.2750 & 6549 & 219 & \underline{0.2313} & 5471 & \underline{171} & 0.7094 & 4859 & \underline{152} & 0.1331 & 2370 & \underline{83} \\
    Kimi-1.5 & 0.3021 & \textbf{5497} & 230 & 0.2354 & 4445 & 175 & \underline{0.7203} & 3913 & 161 & 0.1455 & 2302 & 97 \\
    DAST (reproduce) & 0.2667 & \underline{5531} & \underline{218} & 0.2250 & \textbf{4010} & 183 & 0.7125 & \underline{3539} & 153 & 0.1455 & \textbf{2043} & 90 \\ 
    \textbf{ChainPrune} & \textbf{0.3292} & 5542 & \textbf{178} & \textbf{0.2458} & \underline{4079} & \textbf{134} & \textbf{0.7250} & \textbf{2982} & \textbf{87} & \textbf{0.1734} & \underline{2246} & \textbf{82} \\
    \bottomrule
    \end{tabular}
    \caption{Performance comparison of ChainPrune against baseline methods across multiple reasoning tasks. Bold indicates the best result, underline indicates the second-best result. Token Length and Chain Length are averaged \emph{only over correct answers}. In some cases, ChainPrune appears slightly worse than the best baseline on these metrics; this is because its higher accuracy means that more difficult problems are included in the averaging, which naturally increases the average token and chain length.}
    \label{table:main results}
\end{table*}

\subsection{Direct Preference Optimization}
We adopt Direct Preference Optimization (DPO) combined with a supervised fine-tuning (SFT) loss to optimize the model's reasoning path towards task-specific optimality criteria (conciseness and correctness). The joint objective function is defined as:
\begin{equation}
\begin{aligned}
\mathcal{L}_{\text{DPO+SFT}} &= \mathbb{E}_{(x,y_w,y_l)\sim\mathcal{D}}[-\log\sigma(\beta\log\frac{\pi_\theta(y_w|x)}{\pi_{\text{ref}}(y_w|x)} \\
&- \beta\log\frac{\pi_\theta(y_l|x)}{\pi_{\text{ref}}(y_l|x)})] \\
&+ \lambda\mathbb{E}_{(x,y_w)\sim\mathcal{D}}\left[-\log\pi\theta(y_w|x)\right]
\end{aligned}
\end{equation}
where $\pi_\theta$ is the current policy model, $\pi_{\text{ref}}$ is the reference model, $y_w$ and $y_l$ are the preferred and less-preferred responses, $\beta$ scales the preference term, and $\lambda$ controls the SFT regularization strength. 
The SFT term explicitly maximizes the likelihood of preferred outputs, preventing degeneration into overly terse or incomplete reasoning, while the DPO term implicitly aligns with preference signals. In our later experiments, we provide extensive empirical validation of the necessity of including the NLL term, and in the Appendix, we offer a theoretical analysis based on the \emph{gradient entanglement} phenomenon \cite{yuan2024common}, which explains why margin-based methods like DPO and SimPO may inadvertently synchronize the log-probabilities of chosen and rejected responses.

\section{Experiments}

\subsection{Experiment Setup}
\textbf{Long-COT Models}: We evaluate two long-chain-of-thought models: \textbf{DeepSeek-R1-Distilled-Qwen-7B} and \textbf{DeepSeek-R1-Distilled-Qwen-1.5B}. Both are fully fine-tuned with a learning rate of $5.0 \times 10^{-6}$ using cosine scheduling, 10\% warmup, and trained on 8~$\times$~NVIDIA Tesla A100 GPUs.

\textbf{Datasets}: Training data is based on the cleaned MATH benchmark, containing 9,967 high-quality problem--answer pairs. For each problem, we sample 16 reasoning paths, each capped at 8,192 tokens. Evaluation covers \textbf{MATH500} \cite{hendrycks2021measuring} (500 competition problems), \textbf{AMC23} \cite{amc23} (30 high-school problems), \textbf{AIME24/25} \cite{aime24} (60 olympiad-level problems), and \textbf{LiveCodeBench} \cite{jain2024livecodebench} (programming problems from LeetCode, AtCoder, CodeForces, Aug 2024--Jan 2025; released\_v5). All datasets use a maximum sequence length of 32,768 tokens.

\textbf{Baselines}:  To systematically evaluate the performance of the methods, we compared the following baselines: (1) \textbf{Kimi-1.5} \cite{team2025kimi}: DPO dataset selects correct+shortest samples, rejects correct responses $\geq1.5$ longer or incorrect+longer ones; reproduced faithfully since the model is closed-source and differs in scale. (2) \textbf{DAST} \cite{shen2025dast}: dynamic token budget scoring with ranked contrastive pairs; reproduced from paper as \textbf{DAST (reproduce)} for reliability.  (3) \textbf{Supervised Fine-Tuning (SFT)}: trained on Kimi-1.5 chosen samples.

\textbf{Evaluation Metrics}: We report \textbf{Accuracy}, \textbf{Token Length}, and \textbf{Chain Length} (number of reasoning steps to final answer). 
The latter two are computed only over correctly answered questions. 
Steps are segmented by \verb|\n\n|, following the standard convention in Qwen models. 
Each experiment is run 16 times and averaged.

\begin{table*}[t]
    \centering
    \small
    \setlength{\tabcolsep}{1mm}
    \begin{tabular}{lcccccccc}
    \toprule
        \textbf{Methods} & \textbf{ACC} & \textbf{Tokens} & \textbf{All Tokens} & \textbf{Chains} & \textbf{All Chains} & \textbf{Faulty Reasoning} & \textbf{Invalid Reflection} & \textbf{Redundant Steps}  \\
    \midrule
    Base Model & 0.9240 & 3709 & 4621 & 111 & 146 & 102 & 124 & 132 \\
    SFT & 0.9080 & \textbf{1891} & \underline{2717} & \textbf{52} & \underline{77} & \underline{68} & \underline{86} & \underline{98} \\
    Kimi-1.5 & 0.9040 & 2777 & 4074 & 124 & 138 & 84 & 92 & 102 \\
    DAST & 0.9260 & 2802 & - & - & - & - & - & - \\
    DAST (reproduce) & 0.9180 & 3264 & 3839 & 99 & 123 & 92 & 104 & 107 \\
    \textbf{ChainPrune} & \textbf{0.9300} & \underline{2170} & \textbf{2643} & \underline{62} & \textbf{72} & \textbf{43} & \textbf{47} & \textbf{45} \\
    \bottomrule
    \end{tabular}
    \caption{Reasoning path evaluation using LLM-as-a-Judge on MATH500. All methods are initialized from DeepSeek-R1-Distilled-Qwen-7B. Tokens and Chains report the average token count and reasoning step count only over correct responses. All Tokens and All Chains report the averages computed over all responses, including incorrect ones. Bold values indicate the best result, and underlined values indicate the second-best result.}
    \label{llm as a judge}
\end{table*}

\subsection{Main Results}
\subsubsection{Overall Performance}
We evaluate ChainPrune on two long-COT large reasoning models across four math benchmarks and one code-generation dataset. 
As shown in Table~\ref{table:main results} and \ref{llm as a judge}, ChainPrune consistently outperforms all baselines in accuracy while also reducing both token and chain length. 
These results demonstrate that our framework can improve reasoning efficiency without sacrificing correctness, and in many cases, even enhance it.

\textbf{Accuracy}: Across all datasets, ChainPrune achieves state-of-the-art accuracy, consistently matching or surpassing base model performance. 
This indicates that our pruning and preference optimization steps retain essential reasoning content and can guide the model toward more reliable reasoning patterns. 
By contrast, SFT shows slight degradation due to its lack of explicit efficiency constraints, while DAST (reproduce) and Kimi-1.5 exhibit noticeable accuracy drops.

\textbf{Token Efficiency}: ChainPrune achieves substantial token savings, averaging a 28.1\% reduction over base models while maintaining or improving accuracy. 
For example, on AIME24 it reduces token usage by 22.6\% and on LiveCodeBench by 23.0\%. 
Token averages are computed only over correct answers; since ChainPrune solves more difficult problems, these are included in the average, which can slightly increase its reported token length compared to the best baseline on that metric. 
Nevertheless, ChainPrune remains the best-performing method in most datasets, even under this stricter evaluation.

\textbf{Chain Length}: While Kimi-1.5 explicitly optimizes for the shortest token usage, ChainPrune not only achieves shorter or comparable tokens but also produces substantially shorter reasoning chains. This shows that our method effectively reduces redundancy in both dimensions, rather than sacrificing one for the other. Across all benchmarks, ChainPrune maintains concise chains without harming reasoning quality, further validating the effectiveness of our joint token--step optimization strategy.

\begin{table}
    \centering
    \small
    \setlength{\tabcolsep}{1mm}
    \begin{tabular}{lcccc}
    \toprule
    \multirow{2}{*}{\textbf{Methods}} & \multicolumn{2}{c}{\textbf{AIME24}} & \multicolumn{2}{c}{\textbf{MATH500}} \\
    \cmidrule(lr){2-3} \cmidrule(lr){4-5}
    & ACC & Tokens & ACC & Tokens \\
    \midrule
    Base Model & 0.5479 & 7346 & 0.9240 & 3709 \\
    \midrule
    \textit{Kimi-1.5} w/ DPO & 0.5125 & 6144 & 0.9040 & 2777 \\
    \textit{Kimi-1.5} w/ DPO + NLL Loss & 0.5583 & 6077 & 0.9280 & 2373 \\
    \midrule
    \textit{DAST} w/ SimPO & 0.5375 & 6909 & 0.9180 & 3264 \\
    \textit{DAST} w/ DPO + NLL Loss & 0.5833 & 6188 & 0.9280 & 2572 \\
    \midrule
    \textbf{\textit{ChainPrune}} & \textbf{0.5833} & \textbf{5688} & \textbf{0.9300} & \textbf{2170} \\
    \bottomrule
    \end{tabular}
    \caption{Comparative analysis of different optimization methods under identical training datasets.}
    \label{tab:my_label}
\end{table}

\subsubsection{Evaluate with LLM-as-a-Judge}
\label{llm-as-a-judge}
To assess reasoning quality beyond accuracy, we note that raw step counts measured via ``\verb|\n\n|'' delimiters cannot fully capture whether shorter chains preserve correctness and coherence. Therefore, we adopt a confidence-calibrated \textbf{LLM-as-a-Judge} framework, further verified by human checks for reliability. Specifically, we use three diverse judge models: \textbf{GPT-o1}, \textbf{DeepSeek-R1}, and \textbf{Qwen-QwQ}, to evaluate outputs along three fine-grained dimensions: faulty reasoning, invalid reflection, and Redundant steps. Each judgment is aggregated via majority voting to mitigate individual model biases, and cases with high disagreement are manually reviewed.

As shown in Table~\ref{llm as a judge}, \textbf{ChainPrune} achieves the lowest rates across all dimensions---43 faulty reasoning steps, 47 invalid reflections, and 45 redundant steps---corresponding to reductions of 57.8\%, 62.1\%, and 65.9\% over the base model. In contrast, the base model exhibits severe deficiencies, especially in Redundant Steps (132 steps), highlighting the necessity of systematic reasoning optimization. These results confirm that ChainPrune's structured semantic pruning---merging semantically equivalent nodes via low-cost semantic similarity---substantially cuts invalid reflections and redundancy while preserving core logical integrity. Further evaluation details are provided in the Appendix.

\section{Analysis and Discussion}
\subsection{Comparison of different variants of DPO in Short-Chosen Preference Optimization}
\label{loss_analysis}
We compared DPO \cite{rafailov2023direct} and SimPO \cite{meng2024simpo} in a short-chosen preference optimization setting, where preference pairs $(y_w, y_l)$ were sampled from the reference model and $y_w$ was shorter in token length than $y_l$. However, the narrow distribution of such datasets and the small edit distance between $y_w$ and $y_l$ often lead these methods to adopt biased strategies. Specifically, we observed a reward synchronization collapse effect: instead of improving $y_w$ relative to $y_l$, the model simultaneously lowers the probabilities of both, degrading reasoning performance across benchmarks.

As shown in Table~\ref{tab:my_label}, incorporating an additional NLL loss term into DPO mitigates this issue, consistently maintaining or improving accuracy on mathematical reasoning tasks while reducing token usage. Notably, DPO+NLL improves the accuracy of both Kimi-1.5 and DAST on AIME24 and MATH500. However, in terms of token efficiency, ChainPrune remains superior. This advantage stems from our data construction pipeline: by merging reasoning paths at the semantic level and selecting Pareto-optimal chains, we obtain reasoning paths with fewer tokens, shorter chains, and complete logical structure.

\subsection{Analysis of Reasoning Path Impact and Sample Efficiency}
\begin{figure}
    \centering
    \includegraphics[width=0.9\linewidth]{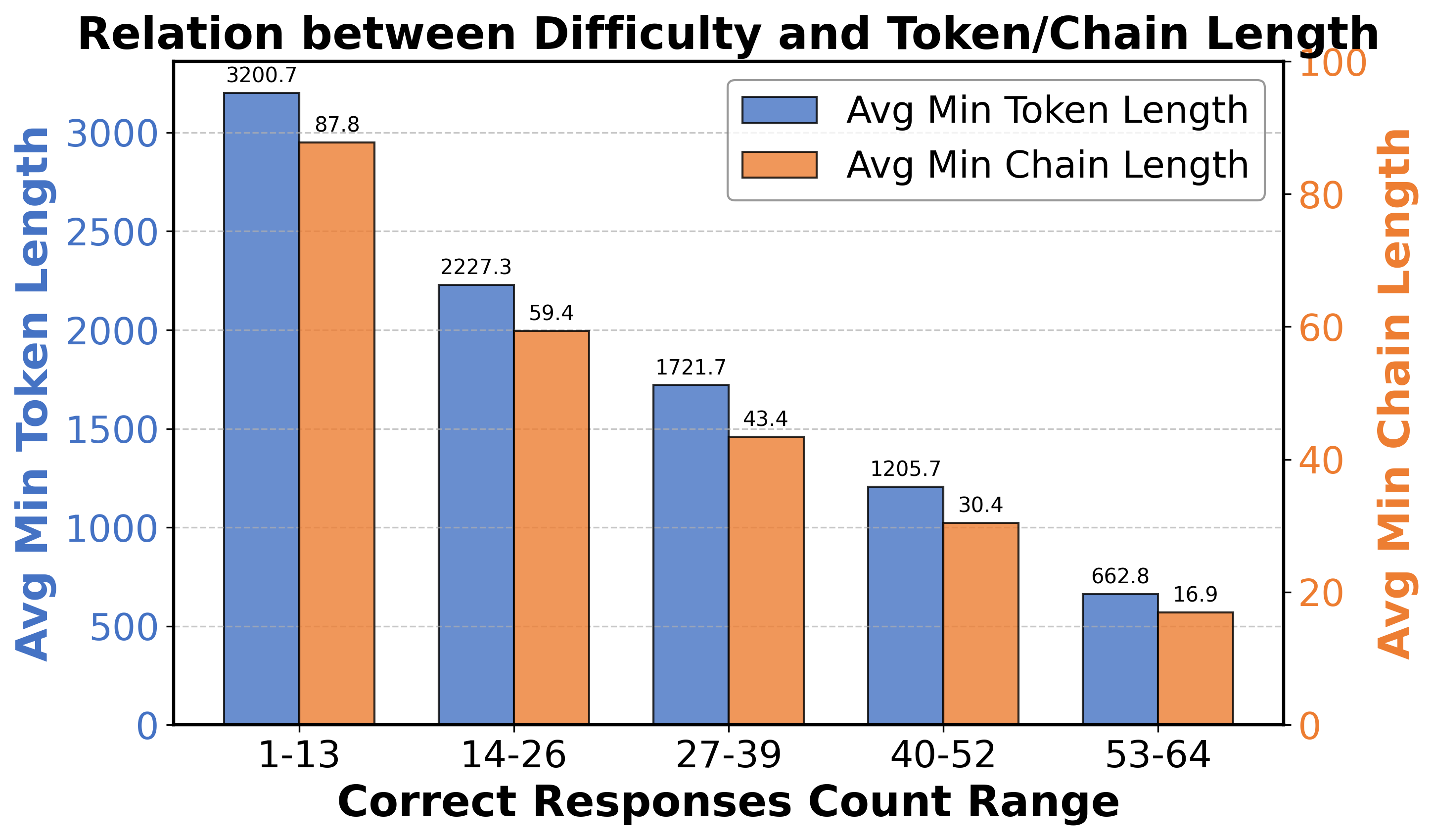}
    \caption{Relationship between difficulty and the token/chain length of the shortest correct response.}
    \label{fig:4a}
\end{figure}

We analyzed the shortest correct responses (by token length) across varying difficulty levels (Figure~\ref{fig:4a}). Token length grows proportionally with problem difficulty---harder problems require longer minimal correct responses. Thus, selecting the shortest correct response naturally adapts optimization to question difficulty.

Ablation studies further support this: shortest responses consistently yielded the most concise outputs, with medium-length responses producing longer outputs, and long responses the longest. This shows a clear linear relationship---shorter chosen samples lead to shorter, equally accurate model outputs. Detailed results are in the Appendix.

\begin{figure}
    \centering
    \includegraphics[width=0.9\linewidth]{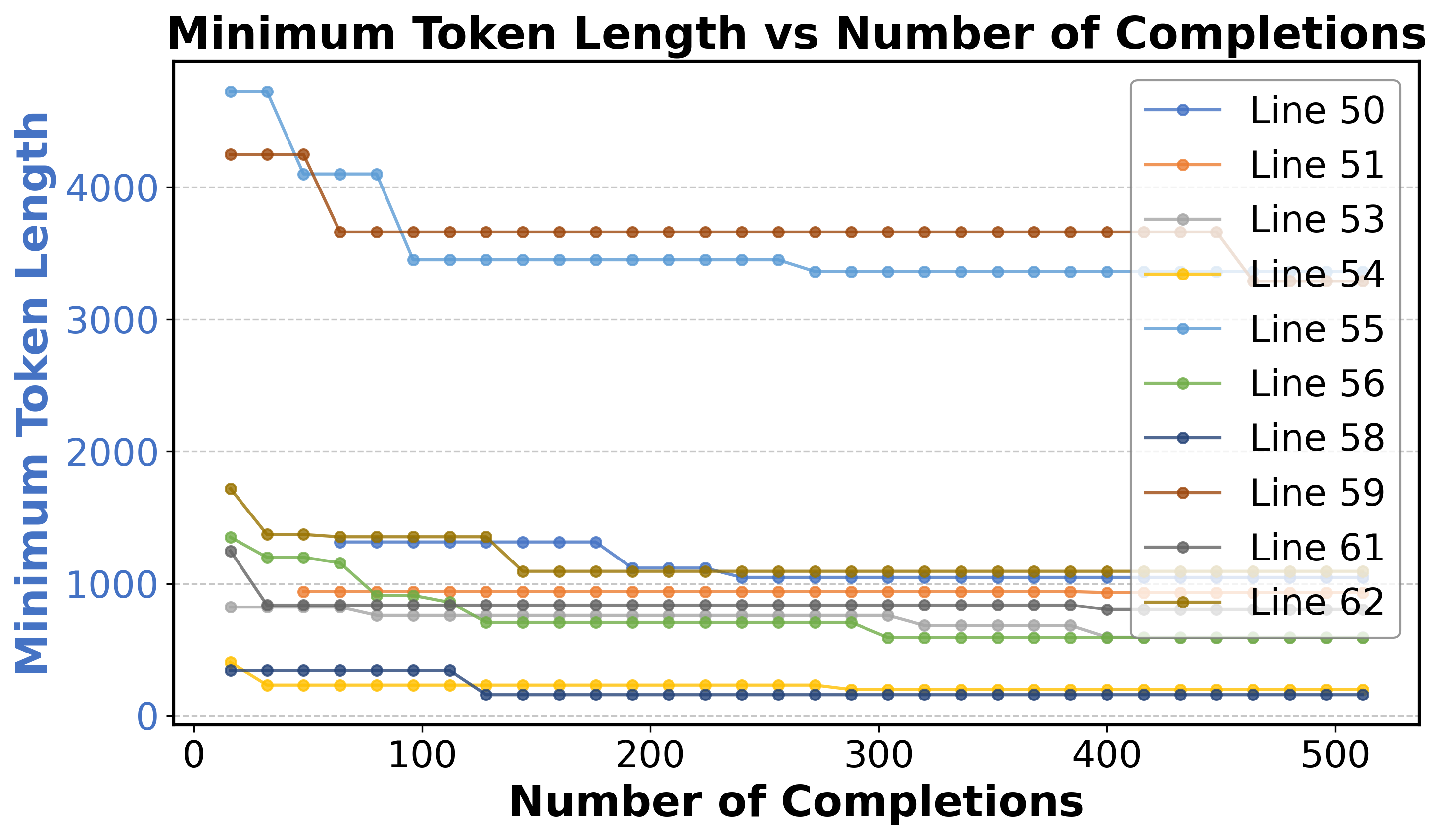}
    \caption{Sample efficiency analysis showing marginal token reduction after extensive sampling.}
    \label{fig:4b}
\end{figure}

However, current RL methods such as DPO and PPO are sample-inefficient: finding sufficiently short correct responses via random sampling is costly. As shown in Figure~\ref{fig:4b}, hundreds of samples yield only marginal token-length gains, making high-quality training data expensive to obtain.

Our method addresses this by merging multiple reasoning paths into a tree structure, enabling efficient generation of high-quality training samples with minimal overhead. As a result, this low-cost approach could be directly integrated into online RL frameworks such as PPO, improving the exploration efficiency for collecting valuable training data.

\section{Related Works}
\textbf{Make Long CoT Short}: Researchers have explored multiple directions to compress reasoning paths while maintaining accuracy. These efforts can be broadly categorized into four key approaches. First, RL with length penalties \cite{team2025kimi, luo2025o1, shen2025dast, hou2025thinkprune, aggarwal2025l1, li2024escape, yang2025think} has emerged as an effective strategy to encourage concise reasoning. Methods like O1-Pruner \cite{luo2025o1} optimize both accuracy and brevity by incorporating length constraints into reward functions, while DAST \cite{shen2025dast} dynamically adjusts reasoning steps based on problem difficulty.
Second, SFT with variable-length CoTs \cite{xia2025tokenskip, yu2024distilling, kang2025c3ot, cui2025stepwise, munkhbat2025self, han2024token, yang2025towards} trains models to generate shorter reasoning paths. TokenSkip \cite{xia2025tokenskip} identifies and skips less critical tokens, while C3oT \cite{kang2025c3ot} leverages LLMs like GPT-4 \cite{achiam2023gpt} to compress reasoning steps. 
Third, prompt-driven efficiency enhancement \cite{renze2024benefits, xu2025chain, chen2024unlocking, lee2025well, aytes2025sketch, chuang2025learning, chuang2025confident} guides models toward concise reasoning without training. Techniques like Concise CoT \cite{renze2024benefits} use simple instructions, while Break the Chain encourages shortcut reasoning.
Finally, latent reasoning \cite{deng2023implicit, shen2025codi, zhang2025lightthinker, cheng2024compressed, liu2024expediting, saunshi2025reasoning, hao2024training} eliminates explicit CoT generation. Implicit-KD \cite{deng2023implicit} distills reasoning into hidden states, while Coconut performs reasoning in a continuous latent space.

\textbf{Guide-based Methods}: To accelerate model inference speed, researchers have proposed guide-based reasoning approaches \cite{yao2023tree, hao2023reasoning, wang2025accelerating, xie2023decomposition}. Gao et al. \cite{gao2024interpretable} pioneered the direct integration of conventional speculative decoding techniques with reasoning methodologies. Building upon this foundation, SEED \cite{wang2024seed} introduced a scheduled speculative decoding framework that coordinates multiple parallel small models through a single shared large model, further enhancing efficiency. The SpecSearch framework \cite{wang2025accelerating} introduces a novel dual-level speculative generator, operating at both coarse-grained thought and fine-grained token levels, achieving accelerated performance without compromising output quality.

\section{Conclusion}
Our work systematically addresses the critical challenge of "pseudo-conciseness" in LLM reasoning optimization, where superficial token reduction fails to eliminate redundant reasoning paths. The proposed ChainPrune method pioneers a semantics-driven compression approach, merging equivalent nodes in reasoning trees to achieve genuine conciseness while preserving logical integrity. Experiments across five reasoning tasks demonstrate ChainPrune's ability to reduce reasoning steps by 26.8\% and tokens by 28.1\% without accuracy degradation, while significantly improving reasoning quality, reducing faulty steps, invalid reflections, and redundant steps by 57.8\%, 62.1\% and 65.9\% respectively compared to the base model. These results establish new standards for evaluating reasoning efficiency, emphasizing semantic coherence over mere token compression.
While the current study focuses on offline optimization, the proposed approach holds strong potential for online RL training frameworks like PPO to enhance exploration efficiency. Due to time and resource constraints, we have not yet implemented this extension, which represents an important direction for future work.

\clearpage
\bibliography{references}

\clearpage
\appendix
\section*{Appendix}
\setcounter{secnumdepth}{1}
\section{Statistical Analysis of Reasoning Path Efficiency}
To verify the hypothesis that token-level optimization in preference learning often overlooks the compactness of reasoning structures, we conducted a comprehensive analysis of all math questions in the training set. For each question, we examined sampled correct reasoning paths to assess the alignment---or misalignment---between minimal token length and minimal chain length. This analysis quantitatively supports our claim in subsection ``Challenges in RL-Optimized Reasoning Path" regarding the issue of pseudo-conciseness.

\begin{figure}[H]
    \centering
    \includegraphics[width=\linewidth]{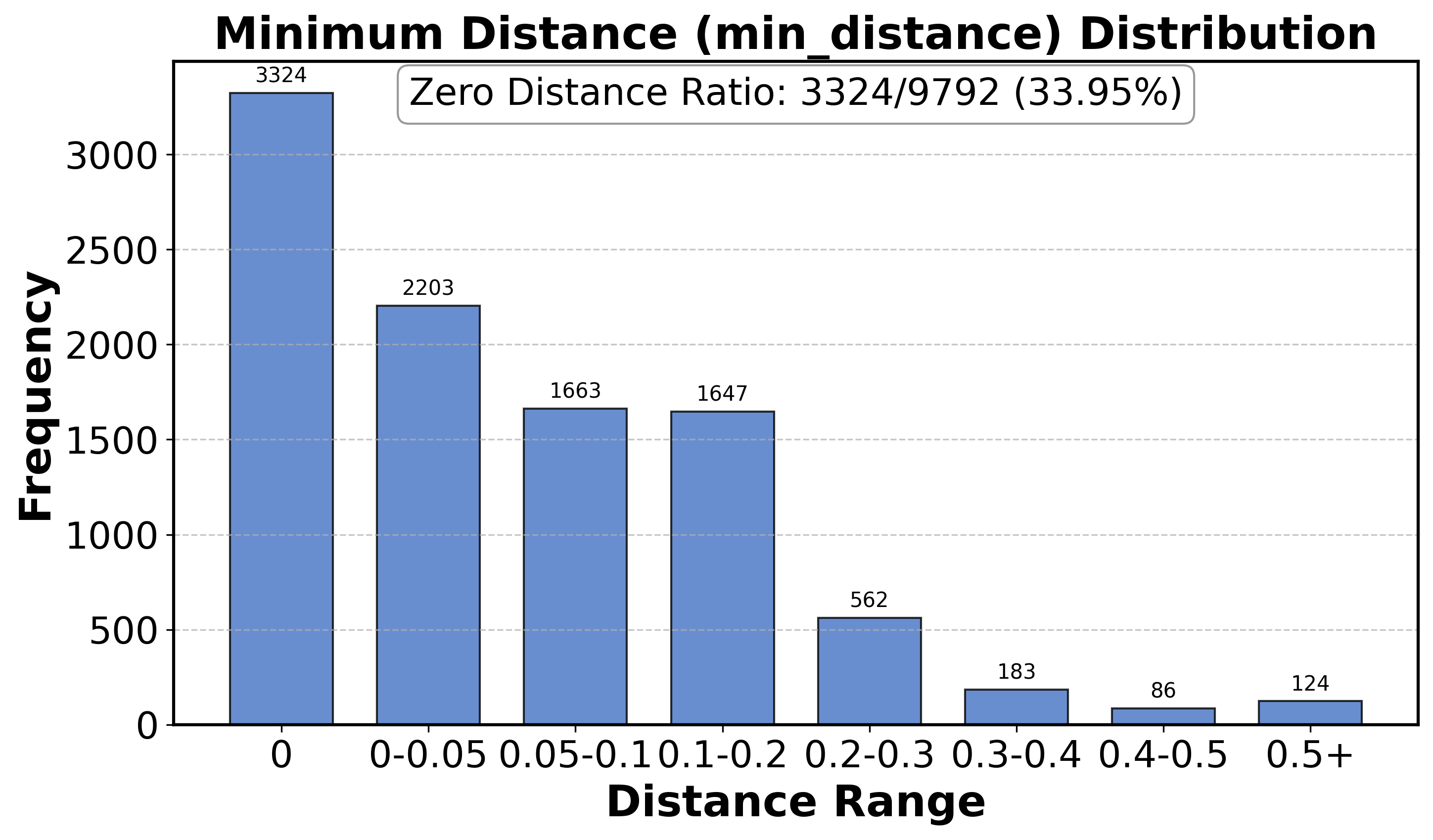}
    \caption{Distribution of \texttt{min\_distance} across all questions. The metric quantifies the normalized discrepancy between the shortest-token and shortest-chain correct reasoning paths. Only 33.95\% of the questions have \texttt{min\_distance} equal to zero, indicating that the same path minimizes both token length and chain length.}
    \label{fig:app-distance}
\end{figure}

\subsection{Evaluation Metrics and Key Findings}
For each question, we sampled multiple correct reasoning paths and computed two key metrics for each path:

\begin{itemize}
  \item \textbf{Token Length}: the total number of output tokens.
  \item \textbf{Chain Length}: the number of reasoning steps in the logical chain.
\end{itemize}
We then ranked all paths by each metric to identify:
\begin{itemize}
  \item $\tau^{\text{min-tok}}$: the path with the shortest token length.
  \item $\tau^{\text{min-step}}$: the path with the fewest reasoning steps.
\end{itemize}

To measure the discrepancy between these two optima, we define a normalized distance metric:
\begin{multline}
\text{min\_distance} = \frac{1}{2}\Bigg(
\frac{|\tau^{\text{min-step}}_{\text{token}} - \text{min}_T|}
{\text{max}_T - \text{min}_T}\\
+\frac{|\tau^{\text{min-tok}}_{\text{step}} - \text{min}_S|}
{\text{max}_S - \text{min}_S}
\Bigg).
\end{multline}
where $\tau^{\text{min-step}}_{\text{token}}$ is the token length of the shortest-step path, and $\tau^{\text{min-tok}}_{\text{step}}$ is the step count of the shortest-token path. $\text{min}_T$, $\text{max}_T$, $\text{min}_S$, and $\text{max}_S$ denote the minimum and maximum values over all paths for the respective metric.
This metric captures how far apart the two optima are in their non-primary dimensions. A value of $\text{min\_distance} = 0$ indicates that both minima are achieved by the same path.

As shown in Figure~\ref{fig:app-distance}, our analysis over the entire training set revealed that only \textbf{33.95\%} of questions have $\text{min\_distance} = 0$, i.e., where $\tau^{\text{min-tok}} = \tau^{\text{min-step}}$. This result highlights a significant misalignment between expression-level and structural conciseness in current optimization schemes.

\begin{figure}[H]
    \centering
    \includegraphics[width=\linewidth]{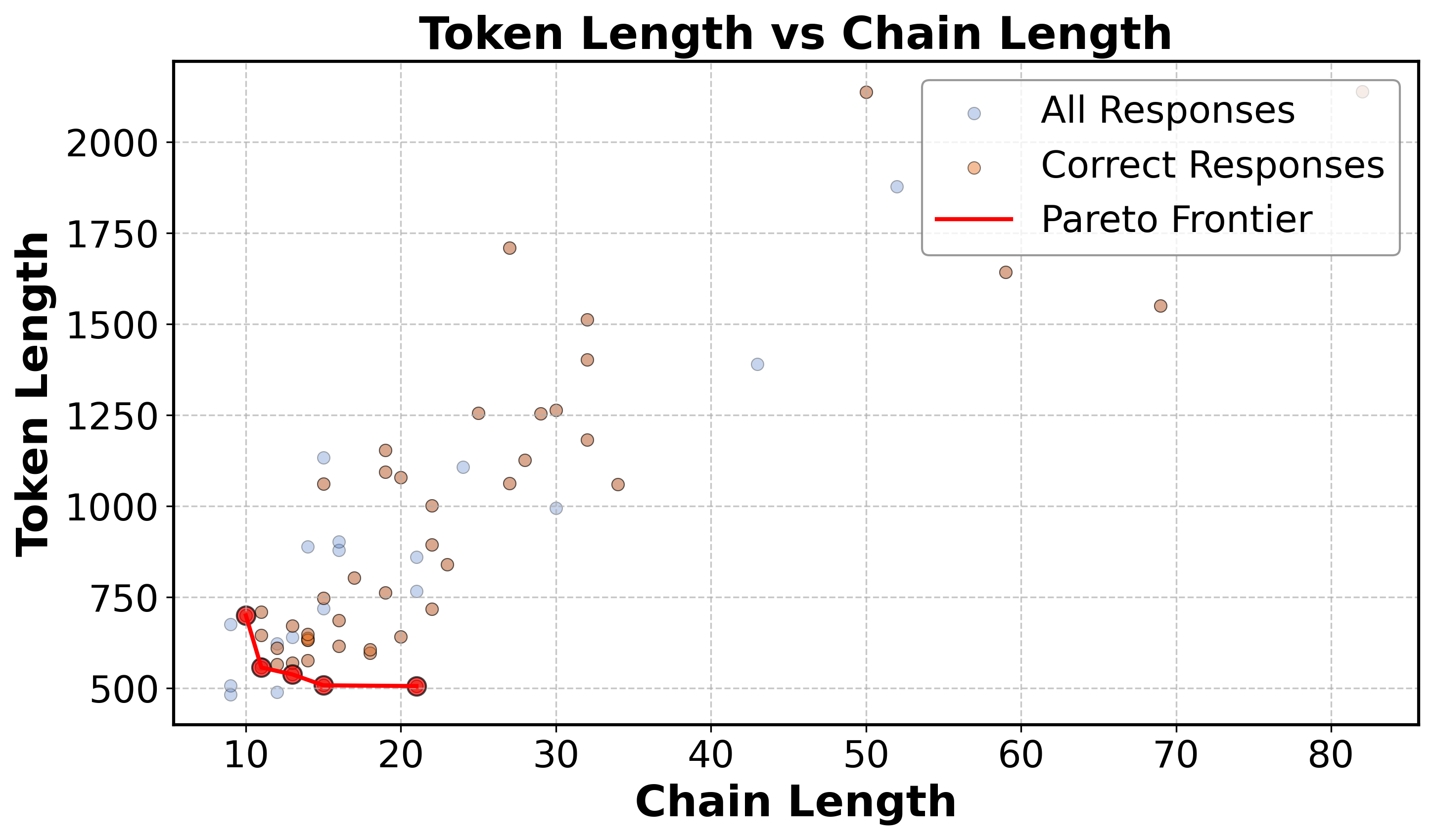}
    \caption{Token length versus chain length for 64 sampled reasoning paths on a representative math question. Each point denotes a unique path. Blue nodes indicate \textbf{wrong responses}, visually distinguishing them from correct ones. No single path lies at the lower-left corner, suggesting that none optimally balances both expression-level and structural conciseness.}
    \label{fig:app-scatter}
\end{figure}

\begin{figure*}[htbp]
    \centering
    \includegraphics[width=\linewidth]{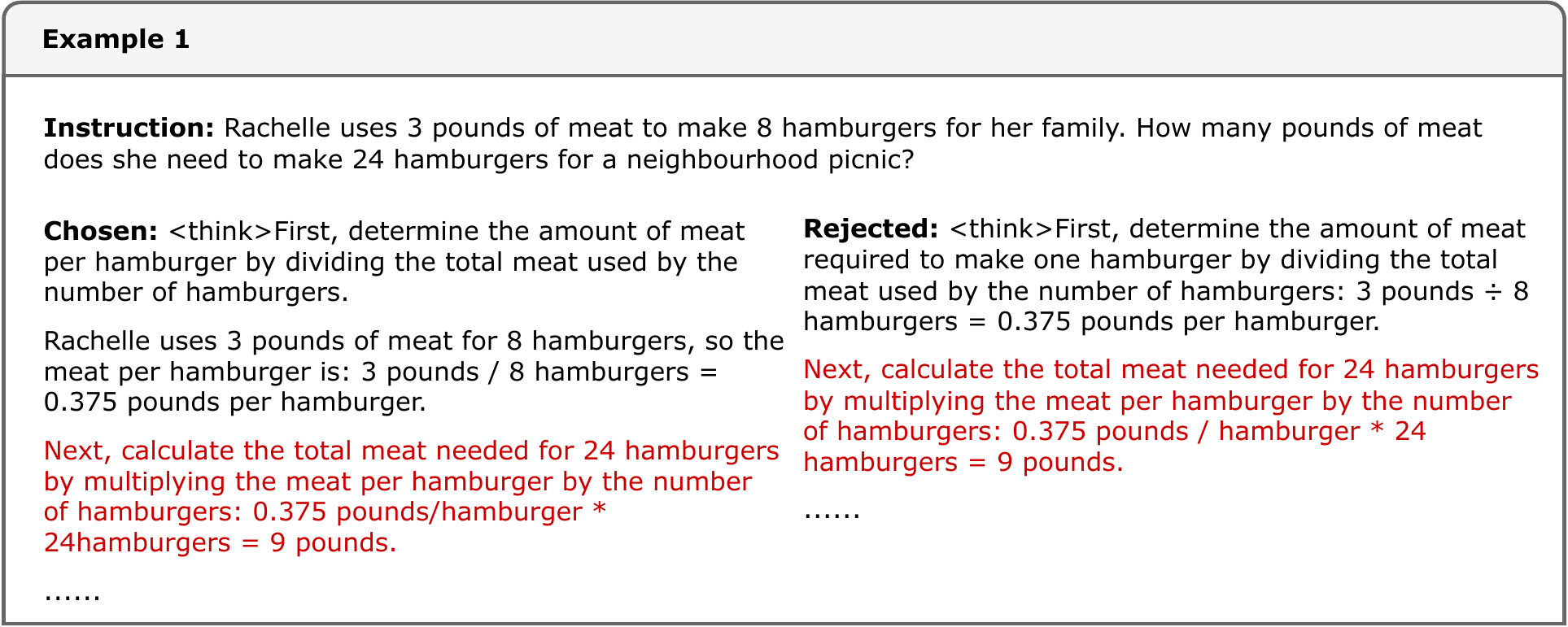}
    \caption{Example 1: A comparison between the chosen and rejected reasoning paths for a proportional reasoning problem. Although both paths share the same semantic steps---computing meat per hamburger and scaling to 24 hamburgers---the rejected path performs these steps earlier.}
    \label{fig:app-example-one}
\end{figure*}

\subsection{Pareto Frontier Visualization}
To further illustrate the misalignment between structural and expression-level conciseness, we randomly selected a representative math question from the training set and sampled 64 reasoning paths. Each path is visualized in a 2D space, with reasoning chain length on the x-axis and token length on the y-axis, as shown in Figure~\ref{fig:app-scatter}.

Each point represents a unique reasoning path. Importantly, no correct path appears in the bottom-left corner of the plot, indicating that joint minimality is not achieved in both metrics. This visual evidence reinforces our central hypothesis: current reward heuristics that aim to minimize token length often fail to promote reasoning paths that are simultaneously compact in structure and concise in expression.

\begin{figure*}
    \centering
    \includegraphics[width=\linewidth]{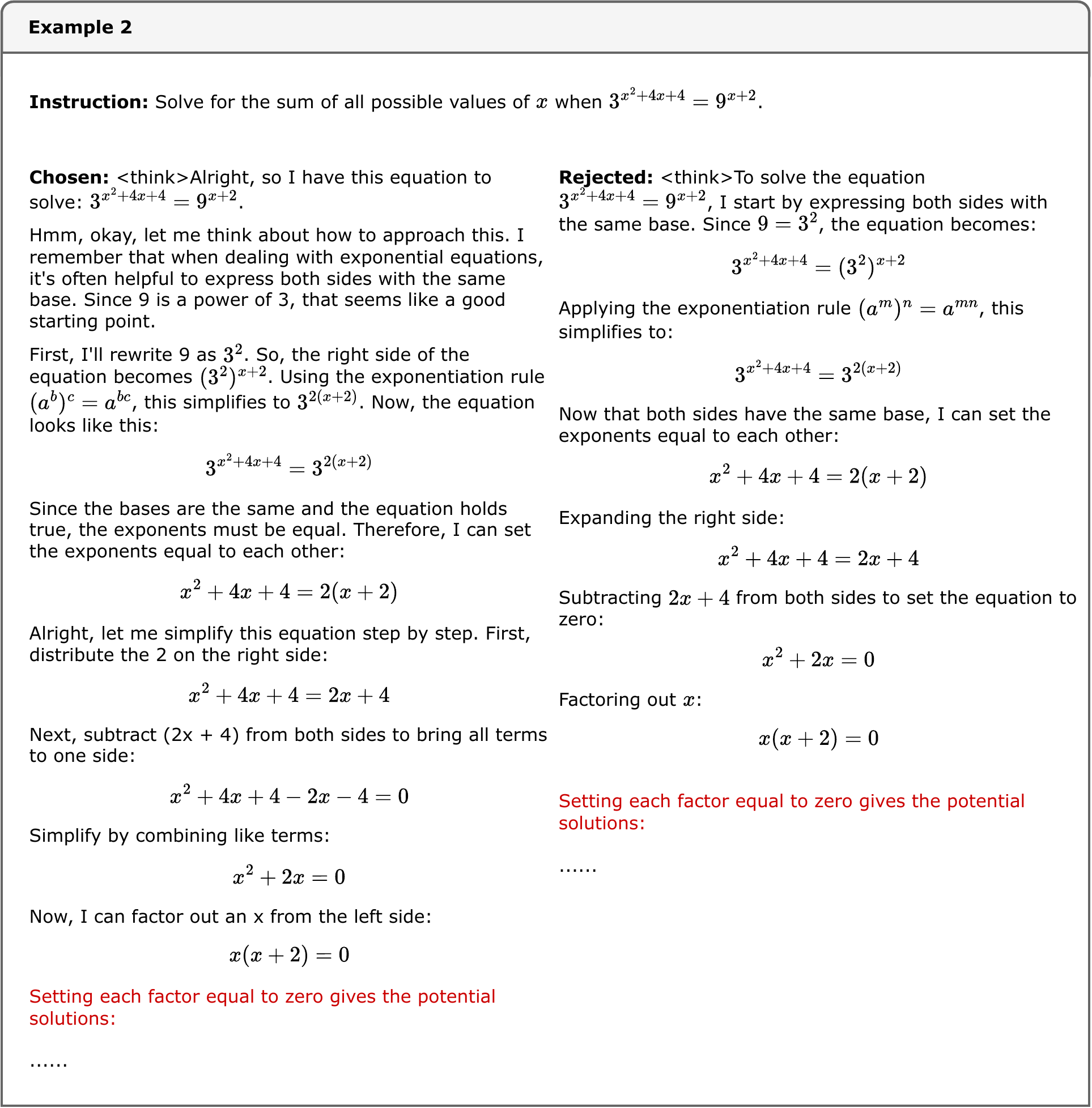}
    \caption{Example 2: A comparison of chosen and rejected reasoning paths for an exponential equation. Both paths apply identical mathematical transformations, including rewriting powers of 9, applying exponent rules, and solving a quadratic equation. However, the rejected path introduces semantically equivalent steps earlier in the trajectory.}
    \label{fig:app-example-two}
\end{figure*}

\section{Examples of Semantically Equivalent Reasoning Steps in Chosen and Rejected Paths}

Our analysis of large reasoning model (LRM) reasoning paths reveals a counterintuitive phenomenon: semantically equivalent intermediate steps coexist in both "chosen" and "rejected" paths, yet systematically emerge at earlier positions in the latter. This suggests a consistent structural bias in the ordering of reasoning content across preference pairs.

In this section, we present concrete examples that illustrate this behavior and trace it to conventional dataset construction methods, which prioritize minimal token length when selecting "chosen" responses. As shown in Figure~\ref{fig:app-example-one} and Figure~\ref{fig:app-example-two}, even when both paths reach the correct final answer and share nearly identical intermediate reasoning steps, the path with more delayed semantic content---and thus lower token count per step---is systematically favored.

This bias results in a counterintuitive preference for syntactically fragmented yet token-efficient reasoning, at the cost of structural clarity and efficiency. 

\section{Multi-Reasoning Paths Merging Algorithm}
\label{sec:app-tree-merging}

In this section, we provide the implementation details of the multi-reasoning path merging algorithm (Algorithm~\ref{alg:app-tree-merging}). The algorithm takes as input a set of $K$ candidate reasoning paths $\{\tau_1,...,\tau_K\}$ generated by a base LLM and constructs a compact, step-level reasoning tree $T$ through hierarchical merging. The goal is to retain all critical reasoning logic while pruning redundancy across paths. The algorithm operates in three main phases:

\begin{itemize}
    \item \textbf{Initialization Phase}: Each path $\tau_i$ is first split into a sequence of reasoning steps $S_i$. The shortest path $S_{\min}$ is selected as the backbone for tree construction, and its steps are used to build an initial linear tree $T$.

    \item \textbf{Tree Construction Phase}: A chain of nodes is created from $S_{\min}$, where each node represents one reasoning step and is attached sequentially to form a linear tree structure. This backbone captures a minimal but coherent reasoning trajectory.

    \item \textbf{Path Merging Phase}: Each of the remaining paths $S_i \ne S_{\min}$ is incrementally merged into the tree. For each step $u$ in $S_i$, we compute its embedding and search for semantically similar nodes in the tree using cosine similarity. If the similarity exceeds a threshold $\theta_{\text{sim}}$, the step is considered a merge candidate. To ensure logical consistency, we further verify that substituting $u$ with the candidate node $v$ does not significantly increase the predictive entropy of the model. Specifically, the entropy difference $\Delta H$ before and after replacement must be within a tolerance threshold $\epsilon$. If both conditions are satisfied, the remaining steps in $S_i$ are attached as a subtree to the matched node. If no valid merge is found, the entire path is appended as a new branch from the root.
\end{itemize}

\begin{algorithm}[htbp]
\caption{Multi-Reasoning Paths Merging Algorithm}
\label{alg:app-tree-merging}
\begin{algorithmic}[1]
\REQUIRE
    \STATE $\textit{paths}$: Set of reasoning paths $\{\tau_1,...,\tau_n\}$ ($<$think$>$ $\tau_i$ $<$/think$>$)
    \STATE $\theta_{\text{sim}}$: similarity threshold for semantic merging
    \STATE $\epsilon$: entropy threshold to ensure logical consistency during merging
\ENSURE 
    \STATE Merged tree structure $T$
    
\STATE \textbf{Initialization:}
\STATE Split each path $P_i$ into step sequence $S_i \gets \textsc{SplitIntoSteps}(\tau_i)$
\STATE Select shortest path $S_{\min} \gets \arg\min_{S_i} |S_i|$
\STATE $\textit{origin\_depth} \gets |S_{\min}|$
\STATE Build initial linear tree $T$ from $S_{\min}$

\STATE \textbf{Build Initial Tree:}
\FOR{$k = 1$ \textbf{to} $|S_{\min}|$}
    \STATE Create node $v_k$ with content $S_{\min}[k]$
    \STATE Attach $v_k$ as child of $v_{k-1}$ (or root if $k = 1$)
\ENDFOR

\STATE \textbf{Process Other Paths:}
\FOR{each path $S_i \neq S_{\min}$}
    \STATE $\textit{merged} \gets \textsc{False}$
    \FOR{$t = 1$ \textbf{to} $|S_i|$}
        \STATE $u \gets S_i[t]$
        \STATE Compute embedding $\mathbf{e}_u$
        \STATE Find candidates $C \gets \{ v\ |\ \text{Sim}(\mathbf{e}_u, \mathbf{e}_v) \ge \theta_{\text{sim}} \}$
        \IF{$C \neq \emptyset$}
            \STATE Sort $C$ by depth (desc), similarity (desc)
            \FOR{each $v^* \in C$}
                \STATE Replace $u$ with $v^*$ in $S_i$ and compute entropy $\Delta H$
                \IF{$\Delta H \le \epsilon$}
                    \STATE Merge: attach remaining steps $S_i[t+1:]$ as subtree to $v^*$
                    \STATE $\textit{merged} \gets \textsc{True}$
                    \STATE \textbf{break}
                \ENDIF
            \ENDFOR
            \IF{$\textit{merged}$}
                \STATE \textbf{break}
            \ENDIF
        \ENDIF
    \ENDFOR
    \IF{not merged}
        \STATE Attach $S_i$ as new branch to root
    \ENDIF
\ENDFOR
\RETURN $T$

\end{algorithmic}
\end{algorithm}

The resulting tree structure $T$ reflects shared reasoning segments as internal nodes, divergent reasoning strategies as branches, and unique answers as leaf nodes. This compact representation allows ChainPrune to preserve logical diversity while eliminating redundant computation.

\section{Theoretical Analysis of different variants of DPO in Short-Chosen Preference Optimization}
\label{sec:app-dpo-analysis}
In this section, we theoretically analyze the decline in reasoning performance observed in the Short-Chosen Preference Optimization scenario for DPO and SimPO. Preference optimization methods like DPO and SimPO, which are margin-based, often lead to synchronized increases or decreases in the log probabilities of the chosen response \( y_w \) and the rejected response \( y_l \), denoted as \( \log\pi_w \) and \( \log\pi_l \). Previous work \cite{yuan2024common} has found that the gradient entanglement effect causes this synchronization. Specifically:  

\textbf{Gradient Entanglement Mechanism:} The optimization objective of DPO aims to widen the gap between the log probability of the chosen response \( \log\pi_w \) and that of the rejected response \( \log\pi_l \), requiring both an increase in \( \log\pi_w \) and a decrease in \( \log\pi_l \). The gradient update direction can be expressed as:  
\begin{equation}
    \Delta\theta \propto d_w\nabla\log\pi_w - d_l\nabla\log\pi_l\nonumber
\end{equation}
where, \( \nabla\log\pi_w \) and \( \nabla\log\pi_l \) are the gradient directions of the log probabilities for \( y_w \) and \( y_l \), respectively. \( d_w \) and \( d_l \) are the derivative weights of the loss function concerning \( \log\pi_w \) and \( \log\pi_l \) (in DPO, \( d_w/d_l = 1 \)).

The synchronized movement of \(\log\pi_w\) and \(\log\pi_l\) occurs when the inner product \(\langle \nabla\log\pi_w, \nabla\log\pi_l \rangle\) between these gradients becomes large relative to their individual norms. This can manifest in two problematic scenarios:
\begin{itemize}[leftmargin=10pt]
    \item Synchronized \textbf{Increase} (\(\log\pi_w \uparrow, \log\pi_l \uparrow\)): Occurs when \(\|\nabla\log\pi_l\|^2 \leq \langle \nabla\log\pi_w, \nabla\log\pi_l \rangle \leq \|\nabla\log\pi_w\|^2\). Here, the gradients are so strongly aligned that both probabilities increase, failing to suppress \(y_l\) adequately. This is particularly detrimental in safety-critical tasks where harmful responses (\(y_l\)) must be actively discouraged.
    \item Synchronized \textbf{Decrease} (\(\log\pi_w \downarrow, \log\pi_l \downarrow\)): Arises when \(\|\nabla\log\pi_w\|^2 \leq \langle \nabla\log\pi_w, \nabla\log\pi_l \rangle \leq \|\nabla\log\pi_l\|^2\). In this case, the model simultaneously forgets both good and bad behaviors, which explains the observed decline in reasoning performance when distilling from high-quality \(y_w\) responses.
\end{itemize}
For DPO (\(\frac{d_w}{d_l}=1\)), the ideal divergence condition (\(\log\pi_w \uparrow, \log\pi_l \downarrow\)) requires:
\begin{equation}
    \langle \nabla\log\pi_w, \nabla\log\pi_l \rangle \leq \min(\|\nabla\log\pi_w\|^2, \|\nabla\log\pi_l\|^2)\nonumber
\end{equation}
Through its length-normalization design (\(\frac{d_w}{d_l} = \frac{|y_l|}{|y_w|}\)), SimPO reformulates this condition as a more lenient inequality:
\begin{align}
\left\langle \frac{\nabla\log\pi_w}{|y_w|}, \frac{\nabla\log\pi_l}{|y_l|} \right\rangle &\leq \left\| \frac{\nabla\log\pi_w}{|y_w|} \right\|^2,\nonumber \\
\left\langle \frac{\nabla\log\pi_w}{|y_w|}, \frac{\nabla\log\pi_l}{|y_l|} \right\rangle &\leq \left\| \frac{\nabla\log\pi_l}{|y_l|} \right\|^2\nonumber
\end{align}

\begin{figure}[ht]
    \centering
    \includegraphics[width=\linewidth]{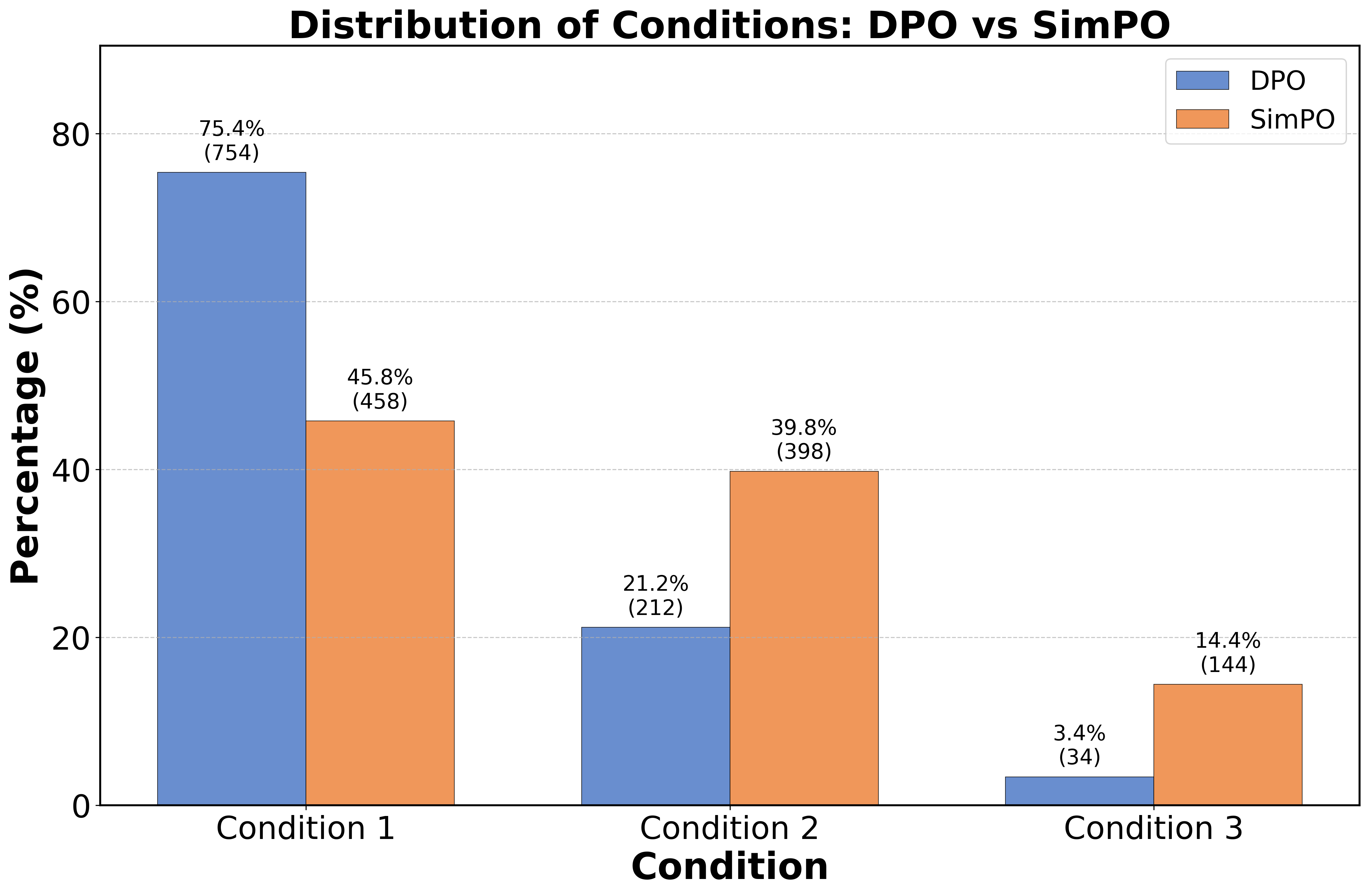}
    \caption{The figure shows three gradient entanglement conditions in DPO/SimPO training: Condition 1 (synchronized decrease: \(\log\pi_w \downarrow, \log\pi_l \downarrow\)) causing forgetting, Condition 2 (synchronized increase: \(\log\pi_w \uparrow, \log\pi_l \uparrow\)) failing to suppress bad outputs, and Condition 3 (ideal divergence: \(\log\pi_w \uparrow, \log\pi_l \downarrow\)); SimPO reduces but doesn't eliminate suboptimal Conditions 1-2 (85.6\% combined).}
    \label{fig:app-gradient-conditions}
\end{figure}

We conduct an empirical study by sampling 1,000 preference pairs from the dataset and tracking the optimization dynamics of \( \log\pi_w \) and \( \log\pi_l \) during training. The results, summarized in a bar chart (Figure~\ref{fig:app-gradient-conditions}), reveal three possible conditions of gradient entanglement: Condition 1 (synchronized decrease: \( \log\pi_w \downarrow, \log\pi_l \downarrow \)), Condition 2 (synchronized increase: \( \log\pi_w \uparrow, \log\pi_l \uparrow \)), and Condition 3 (ideal divergence: \( \log\pi_w \uparrow, \log\pi_l \downarrow \)). Key observations highlight DPO's dominance of undesirable conditions, with Condition 1 (synchronized decrease) accounting for 75.4\% of cases, explaining the reasoning performance decline due to "forgetting" high-quality responses, while Condition 2 (synchronized increase) occurs in 21.2\% of cases, reflecting failure to suppress long responses. The ideal Condition 3 is rare (3.4\%), underscoring DPO's susceptibility to gradient entanglement.  

SimPO's length-normalized design partially mitigates this issue by relaxing the divergence inequality, yet it still exhibits an 85.6\% combined prevalence of Conditions 1 and 2, indicating residual entanglement. This suggests that merely normalizing gradients by response length is insufficient to fully decouple the dynamics of \( \log\pi_w \) and \( \log\pi_l \), leaving room for further improvement in disentangling optimization pathways.

\begin{table*}[htbp]
    \centering
    \small
    \setlength{\tabcolsep}{1mm}
    \begin{tabular}{lcccccccccccc}
    \toprule
    \multirow{2}{*}{\textbf{Methods}} & \multicolumn{3}{c}{\textbf{AIME24}} & \multicolumn{3}{c}{\textbf{AIME25}} & \multicolumn{3}{c}{\textbf{AMC23}} & \multicolumn{3}{c}{\textbf{MATH500}} \\
    \cmidrule(lr){2-4} \cmidrule(lr){5-7} \cmidrule(lr){8-10} \cmidrule(lr){11-13}
    & ACC & Tokens & Chains & ACC & Tokens & Chains & ACC & Tokens & Chains & ACC & Tokens & Chains \\ 
    \midrule
    Longest & 0.5667 & 6576 & 240 & 0.4167 & 5555 & 202 & 0.9094 & 4027 & 137 & 0.9260 & 2427 & 73 \\
    Middle & 0.5833 & 6471 & 235 & 0.4083 & 5441 & 194 & 0.9141 & 3958 & 129 & 0.9200 & 2376 & 67 \\
    \textbf{Shortest} & \textbf{0.5833} & \textbf{5688} & \textbf{225} & \textbf{0.4250} & \textbf{5351} & \textbf{189} & \textbf{0.9125} & \textbf{3771} & \textbf{122} & \textbf{0.9300} & \textbf{2170} & \textbf{62} \\
    \bottomrule
    \end{tabular}
    \caption{Correlation between chosen sample length and optimized model's output length}
    \label{tab:app-length-ablation}
\end{table*}

DPO combined with NLL Loss, breaks the gradient entanglement effect in standard DPO through explicit regularization. The NLL Loss increases the coefficient \(d_w\) for the chosen gradient while leaving the coefficient \(d_l\) for the rejected gradient unaffected, thereby relaxing the condition for increasing the log probability of the chosen response. Simultaneously, the incorporation of the NLL term creates a dynamic balance in the optimization objective: it preserves the contrastive learning advantage of DPO (where the chosen response is favored over the rejected one) while explicitly maximizing the probability of the chosen response, thus avoiding the issue where purely margin-based methods might lead to a decrease in the probability of the chosen response. 

\begin{figure}[htbp]
    \centering
    \includegraphics[width=\linewidth]{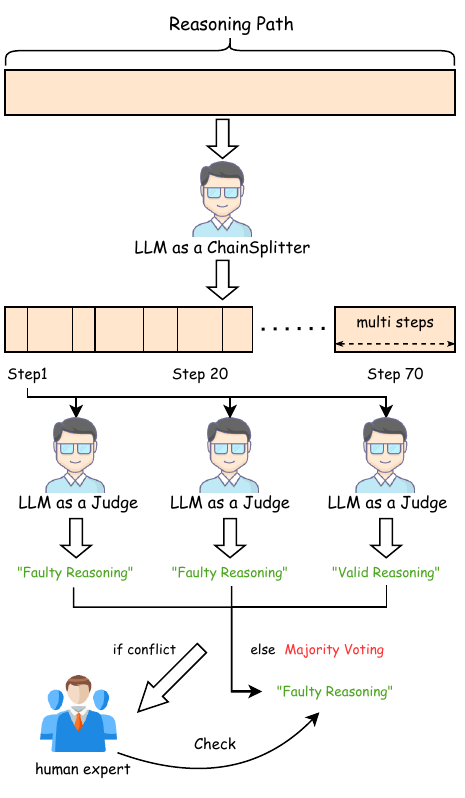}
     \caption{The overall pipeline of the LLM-as-a-Judge framework. The reasoning path is first segmented by GPT-4o acting as a chain splitter. Then, each reasoning step is independently evaluated by three expert models across three categories: faulty reasoning, invalid reflection, and redundant steps. Results are aggregated via majority voting or deferred to human reviewers in cases of high disagreement.}
    \label{fig:app-judge-pipeline}
\end{figure}

\section{LLM-as-a-Judge for Evaluation of Reasoning Model}

We present an improved \textit{LLM-as-a-Judge} framework for evaluating the fine-grained quality of reasoning steps in multi-step inference tasks. Unlike prior approaches that relied on fixed heuristic segmentation (\texttt{\textbackslash n\textbackslash n}) and direct long-context evaluations, our framework enhances precision and scalability through two major innovations: (1) dynamic step segmentation using a large language model as a chain splitter, and (2) a multi-model voting protocol for robust step-level classification.

As shown in Figure~\ref{fig:app-judge-pipeline}, our framework follows a two-stage pipeline consisting of step segmentation and multi-model evaluation.
In the first stage, given a full reasoning path, we employ GPT-4o to segment the path into semantically meaningful steps. This LLM-as-a-Chainsplitter strategy avoids the brittleness of rule-based chunking and ensures that each step reflects a coherent and minimal unit of reasoning. In the second stage, each segmented step is independently evaluated by three expert reasoning models---\textbf{GPT-o1}, \textbf{DeepSeek-R1}, and \textbf{Qwen-QwQ}---along three critical dimensions: \textbf{faulty reasoning}, \textbf{invalid reflection}, and \textbf{redundant steps}. The final label for each step is determined via majority voting; in cases of high inter-model disagreement, a human expert is consulted.

The classification criteria are as follows: (1) \textit{Faulty reasoning} refers to logically incorrect or mathematically invalid steps that jeopardize correctness; (2) \textit{Invalid reflection} denotes ineffective or incoherent meta-cognitive steps that fail to improve or critically assess the reasoning process; (3) \textit{Redundant steps} capture repetitive or non-contributory steps that offer no new information. This categorization schema supports precise error attribution and reflects key weaknesses in reasoning traceability.

\begin{figure}[htbp]
    \centering
    \includegraphics[width=\linewidth]{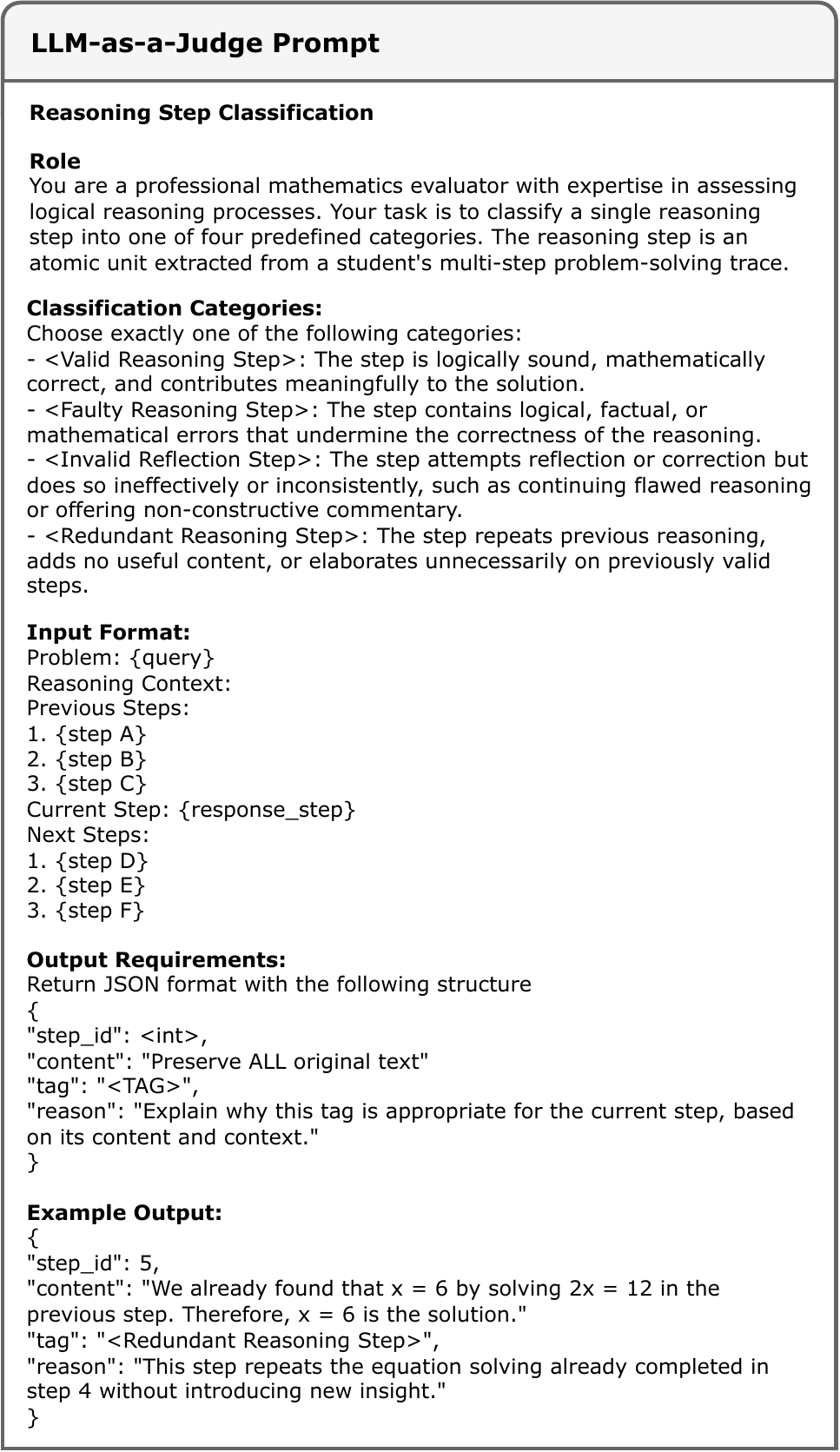}
    \caption{The structured prompt design for the LLM-as-a-Judge step-level classification task. The prompt includes the problem statement, the current reasoning step, and up to three preceding and three following steps as context. The model is instructed to classify the target step into one of four categories and output a JSON object including the classification tag and an explanation.}
    \label{fig:app-judge-prompt}
\end{figure}

To ensure label accuracy and interpretability, we design a structured prompt that provides the current step along with up to three preceding and three following steps as context. As shown in Figure~\ref{fig:app-judge-prompt}, the model is instructed to return a JSON object containing the predicted tag and a justification. This context-aware prompt enables evaluators to distinguish, for example, between truly novel steps and those that are redundant or flawed only in relation to their neighbors.

\section{More Experiment Results}

The ablation experiments in Table~\ref{tab:app-length-ablation} demonstrated a clear trend: shorter chosen samples consistently led to more optimized model outputs, with both reduced token counts and reasoning chains. Specifically, when comparing responses of varying lengths (shortest, medium (5th-shortest reasoning path), and longest (10th-shortest reasoning path)) for the same questions, the shortest samples produced the most concise outputs (e.g., 2,170 tokens vs. 2,427 for the longest in MATH500) while maintaining or even improving accuracy. This suggests that shorter responses not only enhance efficiency by reducing output length but also streamline reasoning chains. The linear correlation held across datasets, with middle-length samples yielding intermediate results, indicating that response brevity is associated with improved computational efficiency.

\end{document}